\documentclass[letterpaper]{article} 
\usepackage[preprint]{aaai2027}  
\usepackage[hyphens]{url}  
\usepackage{graphicx} 
\usepackage{natbib}  
\usepackage{caption} 
\usepackage{algorithm}
\usepackage{algorithmic}

\usepackage{newfloat}
\usepackage{listings}
\DeclareCaptionStyle{ruled}{labelfont=normalfont,labelsep=colon,strut=off} 
\floatstyle{ruled}
\newfloat{listing}{tb}{lst}{}
\floatname{listing}{Listing}

\usepackage{booktabs}

\usepackage{algorithm}
\usepackage{algorithmic}
\usepackage{booktabs}

\usepackage{caption}
\usepackage{subcaption}

\usepackage{amsmath}
\usepackage{amssymb}
\newtheorem{definition}{Definition}

\usepackage{xcolor}

\newif\ifcomments
\commentstrue       

\title{
Diagnosing Temporal Misalignment in Multichannel Time-Series Classification with Minimum Description Length
}

\title{Diagnosing Temporal Misalignment in Multichannel Time-Series Classification via Minimum Description Length}
\author {
    Sebastian Buschjäger\textsuperscript{\rm 1},
    Michael Frichert\textsuperscript{\rm 2},
    Daniel Kuhse\textsuperscript{\rm 2},
    Jian-Jia Chen\textsuperscript{\rm 3}
}
\affiliations {
    \textsuperscript{\rm 1}Lamarr Institute, TU Dortmund University, Germany\\
    \textsuperscript{\rm 2}DAES Group, TU Dortmund University, Germany\\
    \textsuperscript{\rm 2}Cyber-Physical Systems, RWTH Aachen, Germany\\
    \{sebastian.buschjaeger,michael.frichert,daniel.kuhse\}@tu-dortmund.de,
    jian-jia.chen@cps.rwth-aachen.de
}

\begin{document}

\maketitle

\begin{abstract}
Multichannel time-series classification commonly assumes synchronized sensor streams, although latency, clock drift, and preprocessing can introduce relative delays during data collection or after deployment. Existing synchronization solutions are often hardware-specific and difficult to apply retrospectively. Consequently, synchronization problems may remain undetected while classification performance is suboptimal. We introduce a classifier- and label-free diagnostic based on minimum description length (MDL). Our method applies candidate temporal shifts to sensor groups and measures how efficiently one group can be encoded through a representation of the remaining channels. An increased codelength indicates that the shift destroys shared temporal structure, whereas the minimum identifies the alignment most strongly supported by the data. Unlike learned synchronization methods, the diagnostic requires neither retraining nor a trusted aligned reference and can therefore test both training and deployment data for misalignments. Experiments on two controlled synthetic tasks and nine real-world datasets show that the metric exposes alignment structure and can recover accuracy under induced deployment drift. A whole-dataset audit further identifies stable nonzero MDL optima in established benchmarks including FordChallenge, Opportunity, PAMAP2, and UCIActivity, revealing potential systematic offsets that conventional model evaluation does not expose. Our method thus provides a general-purpose tool for detecting, understanding, and correcting temporal misalignment throughout the time-series learning pipeline. Our code is available under \url{https://github.com/sbuschjaeger/mdl-temporal-misalignment}. 
\end{abstract}

\section{Introduction}
\label{sec:introduction}

Multivariate time-series classification is central to many applications in which several sensors observe the same underlying process over time. Examples include human activity recognition from wearable sensors, physiological monitoring, industrial process control, scientific instrumentation, and cyber-physical systems~\cite{Varela:478248,10.1145/3411824,10.1145/2968456.2974012}. In these settings, the individual channels are usually not arbitrary features: they are simultaneous measurements of a shared temporal event. As a result, many classification tasks depend not only on the values observed within each channel, but also on the relative timing among these channels.
This dependence on synchronization is often taken for granted. Most multichannel time-series classifiers assume that the channels are already aligned, or that any remaining timing errors are small enough to be ignored~\cite{aeberhard2011high, kuhse2024sync}. In practice, however, perfect synchronization is difficult to guarantee. Sensors may have different sampling rates, buffering delays, clock drift, preprocessing pipelines, or communication latencies. Even in highly controlled scientific settings, timing and synchronization remain a practical challenge~\cite{Varela:478248, Damerau2018Timing, moscardi:icalepcs2025-webr003}. 

The impact of temporal misalignment is task-dependent. In some cases, a temporal shift has only little effect. If the label depends mostly on one channel, channel-wise magnitudes, or slowly varying trends, then moderate shifts may not substantially change the relevant information. However, if the label is determined by the relative timing between the channels, shifting the timing of a sensor can remove the underlying structure that distinguishes the classes, making misalignment highly destructive. 
This raises a fundamental question:    \emph{Before training a classifier, can we estimate how much a dataset depends on cross-channel temporal alignment?}
We address this question by proposing an alignment sensitivity metric for multichannel time-series classification. Our approach is based on the minimum description length (MDL) principle~\cite{Rissanen1978MDL}. The key intuition is that, in an alignment-sensitive dataset, each sensor group carries some information about the other sensors as well, and hence each group might be used to (crudely) reconstruct the other group's sensor values. If an artificial sensor-shift hurts or improves the reconstruction, then the dataset is alignment-sensitive. If not, information does not seem to be shared across channels. We capture this effect by designing an MDL-based metric that compares an independent-channel representation with a shared-representation code.

We study our resulting metric in both controlled and real-world settings. First, we introduce two synthetic datasets that allow for fine control over temporal misalignment: one dataset in which labels depend on cross-channel synchronization, and one dataset in which the label is robust to shifting a channel. These datasets show that misalignment can be harmless or severe depending on the task structure. Second, we evaluate our metric on nine real-world multivariate time-series datasets from MONSTER~\cite{monster_ts}. For each dataset, we compare the MDL alignment loss against the actual accuracy degradation of four classifiers: a CNN, a ResNet, a late-fusion model, and a magnitude-only CNN. These experiments show that our metric can effectively capture temporal misalignment when it impacts classifier accuracy during deployment and that it can be used to correct it. Moreover, we compare our metric against SyncNet~\cite{ChungZ16a} which corrects misalignment against a stable baseline whereas our metric does not need such a baseline. Finally, we use our metric to audit nominal alignment, finding stable nonzero optima in four datasets suggesting potential misalignment  during data gathering.



\section{Related Work}
Temporal misalignment is a well-known problem for sensor fusion processes~\cite{aeberhard2011high}, with clock drift, different sampling rates, and various communication or processing delays being common sources of fusion errors. The literature contains extensive work on how to \emph{mitigate} temporal misalignment, either by ensuring proper sensor alignment through calibration and synchronization, or by using preprocessing to correct the data~\cite{mair2011spatio,park2020spatiotemporal, qin2018online, nilsson2010time, de2018robust, westenberger2011temporal,Varela:478248, Damerau2018Timing, moscardi:icalepcs2025-webr003}. However, classical synchronization techniques such as cross-correlation and Dynamic Time Warping (DTW) \cite{GonzalesDTWSync, SakoeDTWSpeech, BerndtDTWDataMining} operate directly at the signal level, and thus are difficult to apply to heterogeneous modalities, where signals originate from different physical domains. Originally proposed in the context of audio-video synchronization, SyncNet~\cite{ChungZ16a} uses a two-stream CNN to extract modality-specific embeddings which are used to estimate synchronization offsets across heterogeneous modalities. Most importantly, this approach assumes a synchronized baseline against which signals can be compared. 
A recent parallel line of work focuses on quantifying the \emph{impact} of temporal misalignment itself. Kuhse et al. investigate robustness against temporal misalignment providing a first formal description ~\cite{kuhse2024sync,DBLP:journals/rts/KuhseTHC25,DBLP:conf/ecrts/KuhseGTWBC26}. 
This paper follows this more recent line of work and proposes a new method to detect temporal misalignment without the need for a synchronized baseline.

\section{Alignment Relevance via MDL} 
\label{sec:alignment-relevance-mdl}

Temporal misalignment is crucial whenever sensor groups interact with each other and this interaction is relevant to the task. In an alignment-sensitive dataset, each sensor group may carry information about the other sensors and might therefore be used to reconstruct their values. If an artificial sensor shift hurts or improves reconstruction, then the data seems to contain some shared structure between sensors. If it does not change reconstruction, little structure appears to be shared across channels. We formalize this intuition through the minimum description length (MDL) principle~\cite{Rissanen1978MDL,10.7551/mitpress/4643.001.0001}. 


\subsection{Notation}
Let $ x=(x_1,\ldots,x_C)\in\mathbb R^{C\times T}$ denote a multivariate time-series with $C$ channels and $T\in \mathbb{N}$ discrete time points. Let $y\in\mathcal Y$ be a target class and let $(x,y)\sim\mathcal D$ be a distribution over the pairs $(x,y)$. We are given a finite dataset $D = \{(x^{(i)},y^{(i)})\}_{i=1}^n $ of $n$ observations sampled from $\mathcal D$. Further, let $\mathcal G$ be a collection of non-empty channel groups $G\subseteq\{1,\ldots,C\}.$ In the simplest case, $\mathcal G$ contains all singleton channels. If domain knowledge suggests that several channels should be treated as a synchronized sensor group, $\mathcal G$ can instead contain predefined groups. As an example, consider EEG data obtained from a headset combined with heartbeat and PPG sensors that naturally form two sensor groups. For a group $G\in\mathcal G$, let $x_G$ denote the channels in $G$ and let $x_{\bar G}$ denote all remaining channels. We write $X_G$ to denote the collection of all observations for the channels in $G$. Given a temporal lag $\tau \in \mathbb{N}_0$, we shift only the channels in $G$ relative to the remaining channels. To avoid padding artifacts, we restrict all channels to the overlapping time region after the shift. For a single observation $x$, we write
$$
x_G^\tau(t) := x_G(t+\tau),
\qquad
z_G^\tau(x) := (x_{\bar G},x_G^\tau),
$$
where $z_G^\tau(x) \in\mathbb R^{C\times (T-\tau)}$. Thus, $z_G^0(x)$ denotes the nominally aligned multichannel window. When $\tau > 0$, $z_G^\tau(x)$ denotes the same window after shifting channel group $G$ relative to the remaining channels, and thus, the effective length of $z_G^\tau(x) $ is reduced. For a finite dataset $D$, the corresponding shifted dataset is
$
Z_G^\tau(D)
=
\left\{
z_G^\tau(x^{(i)})
\right\}_{i=1}^n.
$
We omit the labels as our method is label-free. We write $Z$ when the specific group and lag are not important.

\subsection{An MDL-based temporal alignment metric}

Recall that the Minimum Description Length (MDL) principle casts model selection as a data compression problem. Each candidate model defines a way to encode both the model itself and the data not explained by it. By Occam's razor, the preferred model is the one minimizing the resulting total code length. 
We leverage this idea to measure how much a dataset depends on the alignment of (groups of) sensors by fixing the model and introducing artificial lags between sensor groups. 
Formally, we apply the MDL principle to finite sets of windows such as $Z_{G,y}^\tau(D)$. 
Our starting point is the standard two-part MDL view. We write $DL(\cdot)$ for a description length, i.e., the number of bits required to encode an object under a fixed code. For a finite set of multichannel windows $Z$ and a model $M$, this gives
$$
DL(Z,M)
=
DL(M)
+
DL(Z\mid M).
$$
Here, a \emph{model} $M$ is an encoding scheme or \emph{code}, specifying how a window in $Z$ can be encoded as bits. $DL(M)$ is the cost to describe the model, and $DL(Z\mid M)$ the cost to describe the remaining information $Z$ once the model is known.

While the data $Z$ and its shifted variants are fixed once the lag and channel group are chosen, the choice of the model class is central to our metric. As mentioned before, we want to measure how much more difficult reconstructing one group becomes, if a shift occurs. Since a sufficiently capable model may be able to easily reconstruct a signal despite shifts, we have to sufficiently constrain the reconstruction power of the model. Therefore, we map the data into a lower-dimensional embedding space and use this embedding for reconstruction. 

This approach comes with some caveats: A naive choice would be to estimate the representation from the full shifted dataset $Z_G^\tau(D)=(X_{\bar G},X_G^\tau)$. This is problematic because the representation can adapt to the perturbation and continue to describe the shifted group simply because that group helped define it. A short description would then show only that the representation can re-fit the data. We therefore use an anchored shared-representation code. The representation is estimated once from the fixed complement $X_{\bar G}^{\mathrm{ref}}$, common to every lag, and is then used to reconstruct $X_G^\tau$. This makes the metric asymmetric by design: the remaining channels are the reference and the shifted group is the tested part. If $G$ is correctly aligned with the complement, a compact representation extracted from $X_{\bar G}^{\mathrm{ref}}$ should also explain $X_G^\tau$ well. 
If $G$ is shifted away from an aligned lag, the same representation should become less predictive, and the residual should become more expensive to encode.

A sufficiently constrained model (a concrete choice follows) consists of a representation estimator $\phi$, a reconstruction family $f$, reconstruction parameters $\Theta$, and residuals $E$. For a channel group $G$, we write $ S_G = \phi(X_{\bar G}^{\mathrm{ref}})$ and reconstruct each shifted group from this same representation, $\hat X_G^\tau = f_{\Theta_G^\tau}(S_G),$ where the lag-specific parameters $\Theta_G^\tau$ are fitted and part of the codelength. Note that a ``fixed model'' here refers to a fixed code family and anchored representation, not parameters shared across incompatible targets.

The induced shared-representation description length is
\begin{align*}
\ell_{\mathrm{sh}}(Z_G^\tau(D))
=&
DL_{\bar G}(X_{\bar G})
+
DL_S(S_G) \\
&+
DL_\Theta(\Theta_G^\tau)
+
DL_E\!\left(
X_G^\tau
-
f_{\Theta_G^\tau}(S_G)
\right),
\end{align*}
where $DL_{\bar G}$, $DL_S$, $DL_\Theta$, and $DL_E$ are fixed codes for the complement, the anchored representation, the reconstruction parameters, and the residual, respectively. 
As mentioned, we compare this code to an independent-channel baseline that encodes the complement and shifted group separately:
$$
\ell_{\mathrm{ind}}(Z_G^\tau(D))
=
DL_{\bar G}(X_{\bar G})
+
DL_G(X_G^\tau).
$$
The first term appears in both descriptions. Intuitively, the comparison asks whether it is shorter to encode $X_G^\tau$ directly, or to encode an anchored representation from $X_{\bar G}$, the reconstruction parameters, and the residual. This approach is label-free and builds upon the natural patterns occurring in the data to measure alignment dependency. It is straightforward to adopt these definitions to include class information as well, but we did not find substantial improvements doing so and hence focus on the label-free case here. We refer interested readers to the appendix for details. 

We measure compressibility over all windows jointly. A large gain
$
R_G(\tau)
=
\ell_{\mathrm{ind}}\!\left(Z_G^{\tau}\right)
-
\ell_{\mathrm{sh}}\!\left(Z_G^{\tau}\right)
$
indicates the shifted group can be efficiently described from the remaining channels through a shared temporal embedding. To turn this into an alignment cost, we compare the gain for each lag against the best gain under a range of lags as summarized in Definition \ref{def:shared-representation-mdl-alignment-loss}.


\begin{definition}[MDL Alignment Gain]
\label{def:shared-representation-mdl-alignment-loss}
Given a dataset $X$ of $n$ multichannel windows, a collection of channel groups $\mathcal G$, and a maximum lag $\Delta\geq 1$, fix the independent and anchored shared-representation codes $\ell_{\mathrm{ind}}$ and $\ell_{\mathrm{sh}}$. Define the set of evaluated lags as
$
\mathcal T_\Delta
\subseteq
\{\tau\in\mathbb{N}_0 \mid \tau\leq \Delta\}.
$
For a group $G\in\mathcal G$ and lag $\tau\in\mathcal T_\Delta$, let $Z_G^\tau(X)$ denote the windows obtained by shifting group $G$ by lag $\tau$. The gain is
$$
R_G(\tau)
=
\ell_{\mathrm{ind}}\!\left(Z_G^\tau(X)\right)
-
\ell_{\mathrm{sh}}\!\left(Z_G^\tau(X)\right).
$$
Given the maximum gain $
R_G^\star(\Delta)
=
\max_{\tau\in\mathcal T_\Delta}R_G(\tau)
$, the pointwise alignment cost at lag $\tau$ is
$$
Q_G(\tau;\Delta)
=
R_G^\star(\Delta)-R_G(\tau)
$$
measured in bits. The alignment loss of group $G$ is the average raw cost over the non-zero evaluated shifts,
$$
A_G(\Delta)
=
\frac{1}{|\mathcal T_\Delta|-1}
\sum_{\tau\in\mathcal T_\Delta\setminus\{0\}}
Q_G(\tau;\Delta).
$$
The dataset-level alignment loss is the worst group score
$$
A(\Delta)
=
\max_{G\in\mathcal G} A_G(\Delta),
$$
with corresponding group
$
G^\star
=
\arg\max_{G\in\mathcal G} A_G(\Delta).
$
\end{definition}
Note that $Q_G(\tau;\Delta)\geq0$ where smaller is better and $Q_G=0$ corresponds to a best observed lag. 

The score $A_G(\Delta)$ measures how much shared-representation compressibility is typically lost when a group $G$ is shifted expressed in bits, with larger values denoting more information lost. 
As mentioned previously, labels can alternatively be supplied as side information by fitting and coding each class separately and summing the resulting gains. 
We give its definition and an empirical comparison in the appendix. All main-paper results use the label-free definition above. For consistency in the experiments, we use nonnegative lags. However, note this metric can easily be adopted to negative shifts as well. 


\subsection{Correcting Misalignment via the Alignment Gain}
\label{sec:alignment-diagnostics}

The scalar score $A(\Delta)$ measures how sensitive the dataset is to temporal perturbations in the worst group. Plotting the pointwise $Q_G(\tau;\Delta)$ scores for different $\tau$ contains additional information. In particular, the best observed lag
$
\tau_G^\star
=
\arg\min_{\tau\in\mathcal T_\Delta} Q_G(\tau;\Delta)
$
indicates which relative shift gives the most compact shared-representation description of group $G$ from the remaining channels.

This makes the metric useful for diagnosing misalignment without training a classifier. If $\tau_G^\star=0$, the nominal alignment is best within the evaluated range. If $\tau_G^\star>0$, a delayed version of $G$ gives a shorter shared description and is a candidate correction. In practice, $\tau_G^\star$ should be interpreted together with the raw cost at nominal alignment $Q_G(0;\Delta)$. If this cost is close to zero, then the nominal alignment is nearly as good as the best observed lag, even if the optimum is not unique. Similarly, if there is no clear trend in the data, then interpretation should be cautious. A non-zero $\tau_G^\star$ may reflect noise, periodicity, or a flat alignment curve rather than a meaningful synchronization error. Conversely, a large $Q_G(0;\Delta)$ together with a stable non-zero $\tau_G^\star$ is evidence for a systematic delay already present in the data. In this case, closer inspection of data and its measuring process is in order. 

\subsection{Code Instantiations}
\label{sec:shared-representation-codes}

The above definition leaves open how the anchored representation $\phi$, reconstruction family $f_\Theta$, and component codes are instantiated. We state three requirements. First, the representation must be anchored in $X_{\bar G}^{\mathrm{ref}}$ and not estimated from $X_G^\tau$. Second, the model must be sufficiently simple or it could reconstruct each channel independently. Third, the code family and representation must be fixed before evaluating lags.

For the remainder of the paper, we use deliberately simple choices that work well across our experiments. Details are in the appendix and supplementary code. We Z-normalize each channel and quantize it at 64 units per training standard deviation. To exploit temporal smoothness, we apply a first-order delta transform to each window and treat its first value and subsequent differences as separate integer streams, coded with conditional empirical entropy~\cite{CoverThomas2006}. For $\phi$, we compute rank-one PCA once on the fixed, unshifted complement. We stack samples and time points into $ X_{\bar G}^{\mathrm{ref}}\in\mathbb R^{N\times C_{\bar G}}$ and compute $S_G=\phi_{\mathrm{PCA}}(X_{\bar G}^{\mathrm{ref}})\in\mathbb R^{N\times 1}.$ With one complement channel, this reduces to that channel itself. For more channels, $S_G$ is its dominant one-dimensional component. The reconstruction family is linear:
$
\hat X_G^\tau
=
S_G W_G^\top+\mathbf 1\mu_G^\top,
$
where $\Theta_G^\tau=(W_G,\mu_G)$ are fitted for the shifted group at lag $\tau$, subject to $W_G\geq0$, while the offset $\mu_G$ remains unrestricted. The non-negative weights prevent an anti-phase signal from being treated as aligned merely by negating the shared component. We fix the otherwise arbitrary sign of each PCA component by orienting its largest-magnitude loading positively. Note, that this PCA basis is deterministic given the decoded complement, and it has a deterministic sign. An explicit lag-invariant basis cost would cancel from $Q_G$, so it is not included in the codelength computation. Similar to before, we also encode the anchored representation $S_G$ with the delta entropy code. The reconstruction residual $X_G^\tau-\hat X_G^\tau$ is rounded to integer symbols and encoded using empirical entropy without delta encoding. Each reconstruction slope and intercept is assigned 16 bits. 

\section{Experimental Analysis}
\label{sec:experiments}

Recall that our main goal is to characterize temporal misalignment in data \emph{without} a clean baseline. We therefore test whether our metric recovers known alignment structure in synthetic benchmarks and if it supports lag selection in real-world data. Our experiments address three questions:
\begin{enumerate}
    \item On controlled synthetic data, does the metric expose known alignment structure and distinguish an alignment-sensitive task from a robust one?
    \item On real data, can the label-free lag curve identify misalignment in the data?
    \item On real data, can the label-free lag curve identify and correct an induced deployment lag?
\end{enumerate}

We evaluate these questions on two synthetic datasets and nine real-world multivariate time-series datasets from MONSTER~\cite{monster_ts} depicted in Table \ref{tab:datasets}. The real-world datasets cover applications from human activity recognition to physiological sensing and workload estimation. We use MONSTER's precomputed five-fold splits and five stratified folds for the synthetic data. During pre-experiments, we selected dataset-specific maximum lags from $\Delta=8$ for FordChallenge to $\Delta=80$ for STEW, and $\Delta=1500$ for the synthetic datasets. Please note, that while our metric also supports nonnegative lags, we focus on positive lags only here because SyncNet~\cite{ChungZ16a} only supports positive lags. We train four classifiers: a CNN, a ResNet~\cite{DBLP:journals/corr/HeZRS15}, a Late-Fusion CNN~\cite{Jose2020CNNFusion} that uses independent convolutional branches for each sensor group and a magnitude CNN~\cite{Zhou2018Magnitude} that is trained on the log-magnitude spectrum of the Fourier transform of each channel. Inputs are z-normalized per channel using statistics computed on the training split only. We reserve a stratified 15\% of each training fold for validation and early stopping with at most 100 epochs with a batch size of 256. We optimize cross-entropy using Adam with a learning rate $10^{-4}$, halve the rate after three epochs without validation-loss improvement, and stop after ten such epochs. Additional results can be found in the appendix. Our experiments required ${\approx}29$ GPU-hours on NVIDIA A100 GPUs (25 h for training training, 4 h for the evaluation). The code for these experiments is available under \url{https://github.com/sbuschjaeger/mdl-temporal-misalignment}.

\begin{table}
\centering
\caption{Datasets used in the experiments. The table reports the number of windows $n$, window length $T$, channels $C$, channel groups $|\mathcal G|$, and classes $|\mathcal Y |$. Real-world datasets (top group) are part of MONSTER \cite{monster_ts}.}
\label{tab:datasets}
\begin{tabular}{lrrrrr}
\toprule
Dataset & $n$ & $T$ & $C$ & $|\mathcal G|$ & $|\mathcal Y |$ \\
\midrule
CrowdSourced & 12,289 & 256 & 14 & 14 & 2 \\
DreamerA & 170,246 & 256 & 14 & 14 & 2 \\
DreamerV & 170,246 & 256 & 14 & 14 & 2 \\
STEW & 28,512 & 256 & 14 & 14 & 2 \\
Opportunity & 17,386 & 100 & 113 & 8 & 5 \\
PAMAP2 & 38,856 & 100 & 52 & 4 & 12 \\
Skoda & 14,117 & 100 & 60 & 20 & 11 \\
UCIActivity & 10{,}299 & 128 & 9 & 2 & 6 \\
FordChallenge & 36,257 & 40 & 30 & 3 & 2 \\
\midrule
SyncSine-Sensitive & 3,000 & 100 & 2 & 2 & 2\\
SyncSine-Robust & 3,000 & 100 & 2 & 2 & 2 \\
\bottomrule
\end{tabular}
\end{table}

\subsection{Synthetic Benchmarks}
\label{sec:synthetic-data}

In this experiment we study the behavior of our metric on a synthetic benchmark. To do so, we introduce two synthetic datasets, one that is sensitive and one that is robust to misalignment. Both datasets contain two channels and binary labels. The first dataset is designed to be alignment-sensitive: the label depends on a shared temporal pattern that is present in both channels. The second dataset is designed to be alignment-robust: the label depends only on one channel, while the other channel is a nuisance signal. In both cases, we misalign only the second channel.

For the alignment-sensitive dataset, both channels $c\in\{0,1\}$ contain the same latent sine wave with independent channel offsets $b_c$ and sample noise $\epsilon_c(t)$:
$$
x_c(t)=\sin(t/r+\varphi)+b_c+\epsilon_c(t)
$$
The label depends on the sum of the two aligned channels over the observed window:
$$
y
=
\mathbb I\left[
\frac{1}{T}
\sum_{j=1}^{T}
\left(
x_1(t_j)+x_2(t_j)
\right)
>
1
\right].
$$
where $b_c\sim\mathcal N(0,0.7^2)$ and $\epsilon_c(t)\sim\mathcal N(0,0.05^2)$ independently for each channel. The independent offsets make neither channel alone sufficient for the sum-based label, while the shared oscillation preserves a well-defined cross-channel alignment. Shifting either channel changes the cross-channel relation that defines the label.

\begin{figure}[h]
     \centering
     \begin{subfigure}[b]{\linewidth}
         \centering
	         \includegraphics[width=\textwidth]{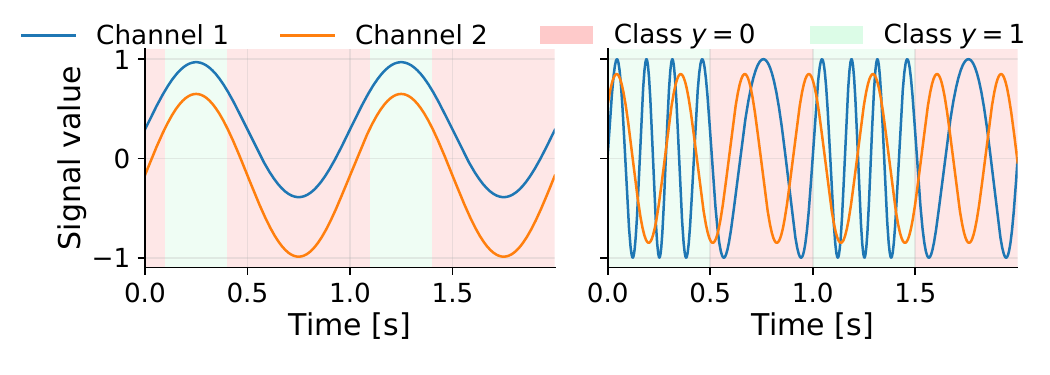}
     \end{subfigure}
     \begin{subfigure}[b]{\linewidth}
         \centering
         \includegraphics[width=\textwidth]{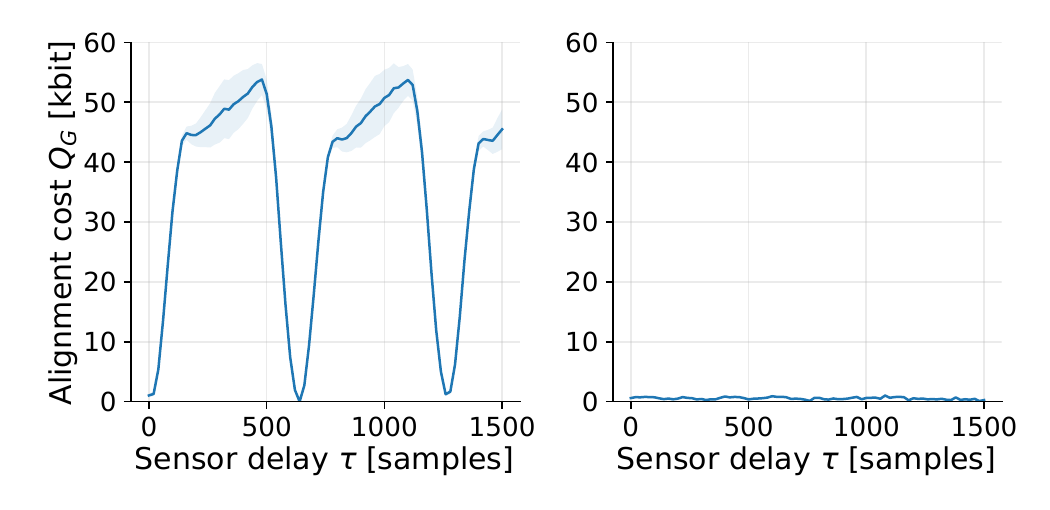}
     \end{subfigure}
     \begin{subfigure}[b]{\linewidth}
         \centering
         \includegraphics[width=\textwidth]{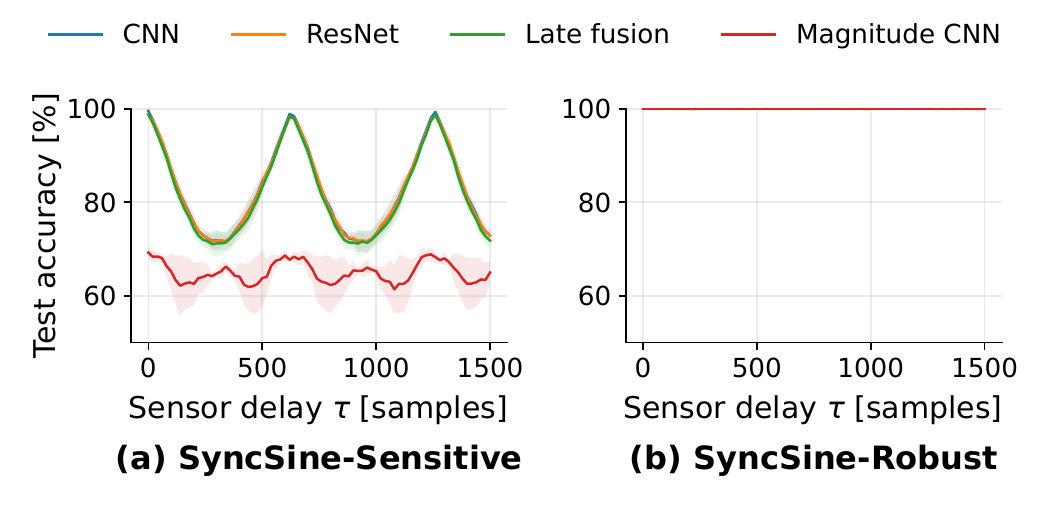}
     \end{subfigure}
    \caption{Results on the synthetic experiments. The left column shows results for the sensitive datasets, and the right column shows results for the robust one. The first row depicts an example of both datasets, where class labels are depicted as shaded regions. The second row shows the alignment costs according to our metric, and the last row shows the accuracy for all four classifiers with the shaded region depicting the standard deviation over 5 cross-validation folds.}
    \label{fig:synthetic_experiments}
\end{figure}

For the alignment-robust dataset, the two channels are generated independently as noisy sinusoids:
$$
x_c(t)=a_c\sin(2\pi f_ct+\varphi_c)+b_c+\epsilon_c(t).
$$
The binary label selects the frequency band of the first channel: $f_1\sim\mathrm{Unif}(2,4.5)$ for class zero and $f_1\sim\mathrm{Unif}(5.5,8)$ for class one. The second-channel frequency is an independent nuisance variable sampled uniformly from $[2,8]$. Amplitudes are sampled from $[0.8,1.2]$, phases from $[0,2\pi]$, offsets from $\mathcal N(0,0.2^2)$, and sample noise from $\mathcal N(0,0.1^2)$. An example can be found in Figure \ref{fig:synthetic_experiments}. Each dataset consists of $3~000$ windows. The observed window length is $T=100$. To evaluate shifts without padding, we generate a longer signal of length $T+\Delta_{\max}$ 
and then restrict shifted windows to the overlapping region. Time points are given by $t_j = j\cdot 0.01.$ For the sensitive dataset, $r=1$ and phases are sampled uniformly from $[0,2\pi]$. For both synthetic datasets, the candidate channel groups are the two singleton groups $\{1\}$ and $\{2\}$. The class distribution is 50:50 for the alignment-robust dataset and $\approx$70:30 (class~0:class~1) for the alignment-sensitive dataset.
We evaluate non-negative lags from 0 to 1500 samples in steps of 20. 

Figure \ref{fig:synthetic_experiments} shows the result, with the sensitive datasets in the left column and the robust one on the right. The first row shows an example of each, with class labels depicted as shaded regions and noise omitted for visual clarity. Clearly, for the sensitive data, the label depends on the sum of both signals, whereas for the robust one, only on the frequency of Channel 1. The second row shows the alignment costs of our metric in bits. One can see a clear increase in the number of bits required for encoding when misalignment is introduced into the data. Since the sine wave data has a periodicity of $2\pi$, temporal patterns recur at roughly $\tau = 618$ samples, where the channels align again and the bit length decreases drastically. In stark contrast, the alignment costs of the robust dataset do not change with different lags and stay close to 0. The accuracy in the last row shows a similar pattern: CNN, ResNet and Late Fusion all achieve 100\% accuracy on aligned data, but any lag introduced quickly destroys it to just above 70 \%, the worst-case accuracy considering the class distribution of $\approx 70:30$. Magnitude CNN is more robust, but weaker performance between 60-70\% accuracy. On the robust dataset, however, all methods achieve 100\% accuracy.

\subsection{Post-Hoc Synchronization of Real-World Datasets}


Data may already contain a systematic offset during training. A classifier trained under these circumstances can appear robust because it may adapt to the offset or never exploit the disrupted cross-channel interactions which can lead to overall worse performance. Our metric can question the original alignment without training a classifier and without requiring an already synchronized baseline. Hence, it can serve as a debugging method for the data-gathering process. We investigate the datasets in Table \ref{tab:datasets} for potential misalignment by computing our MDL metric on the entire dataset and selecting the group with the largest alignment-loss score $A_G(\Delta)$ and reporting its optimal lag $\tau^*_G$. In addition, we compute the lag for each training fold to determine whether our metric is stable or not.
\begin{table}
  \centering
  \small
  \caption{Sensor groups with the largest misalignment in each dataset. $^*$Reached the maximum lag tested, so a larger lag misalignment might be present in the data.}
    \label{tab:global-alignment}
    \begin{tabular}{@{}llrrr@{}}
    \toprule
    Dataset & Group & \begin{tabular}[c]{@{}r@{}}Global \\ $\tau_G^\star$\end{tabular} & \begin{tabular}[c]{@{}r@{}}Training-fold \\ $\tau_G^\star$\end{tabular} & \begin{tabular}[c]{@{}r@{}}$Q_G(0)$ \\ {[}kbit{]}\end{tabular} \\ \midrule
    FordChallenge & env & 6 & 4--8 & 248.5    \\           
    Opportunity & r shoe & 15 & 8--15 & 32.2  \\           
    PAMAP2 & ankle & 20 & 8--20 & 807.4       \\           
    UCIActivity & gyro & 56 & 56 & 389.4  \\           
    Skoda & s05 & 0 & 0--0 & 0.0              \\           
    CrowdSourced & O2 & 0 & 0--0 & 0.0        \\           
    DREAMERA & ch 12 & 0 & 0--0 & 0.0         \\           
    DREAMERV & ch 12 & 0 & 0--0 & 0.0         \\           
    STEW & ch 10 & 0 & 0--0 & 0.0             \\ \bottomrule
    \end{tabular}
\end{table}
Table \ref{tab:global-alignment} shows the results for alignment results. The whole-dataset test identifies nonzero MDL optima for FordChallenge, Opportunity, PAMAP2, and UCIActivity. The corresponding training-fold optima are consistently nonzero, with reasonably stable ranges, which suggest that these datasets are misaligned and they should be corrected.
The remaining five datasets have global and fold-wise optima at zero. In a follow-up experiment, we trained different classifiers after manual correction of the four candidate datasets. However, we found that this does not consistently improve accuracy (see appendix for details). Opportunity is the only dataset with a notable positive result: A ResNet improves by approximately $1.0$ percentage point. We find this seemingly negative result is not necessarily contradictory: a systematic offset can be a reproducible property of the collected data without harming a classifier trained and evaluated under the same convention. Such a classifier may adapt to the offset, while retrospective shifting cannot recover discarded boundary samples, change the temporal semantics of labels, or reconstruct information lost during acquisition. The already processed, windowed data also makes proper realignment difficult in hindsight. Put differently, we cannot sufficiently correct these datasets now that they are already published, since an intervention during data gathering is required. We therefore view this experiment as supporting the idea that the MDL metric is a tool for auditing and debugging synchronization in the data pipeline while data is gathered, rather than as a procedure that is guaranteed to improve an already collected classification benchmark.

\subsection{Accuracy Correction on Real-World Datasets}

\begin{figure*}[h]
    \centering
    \includegraphics[width=0.76\linewidth]{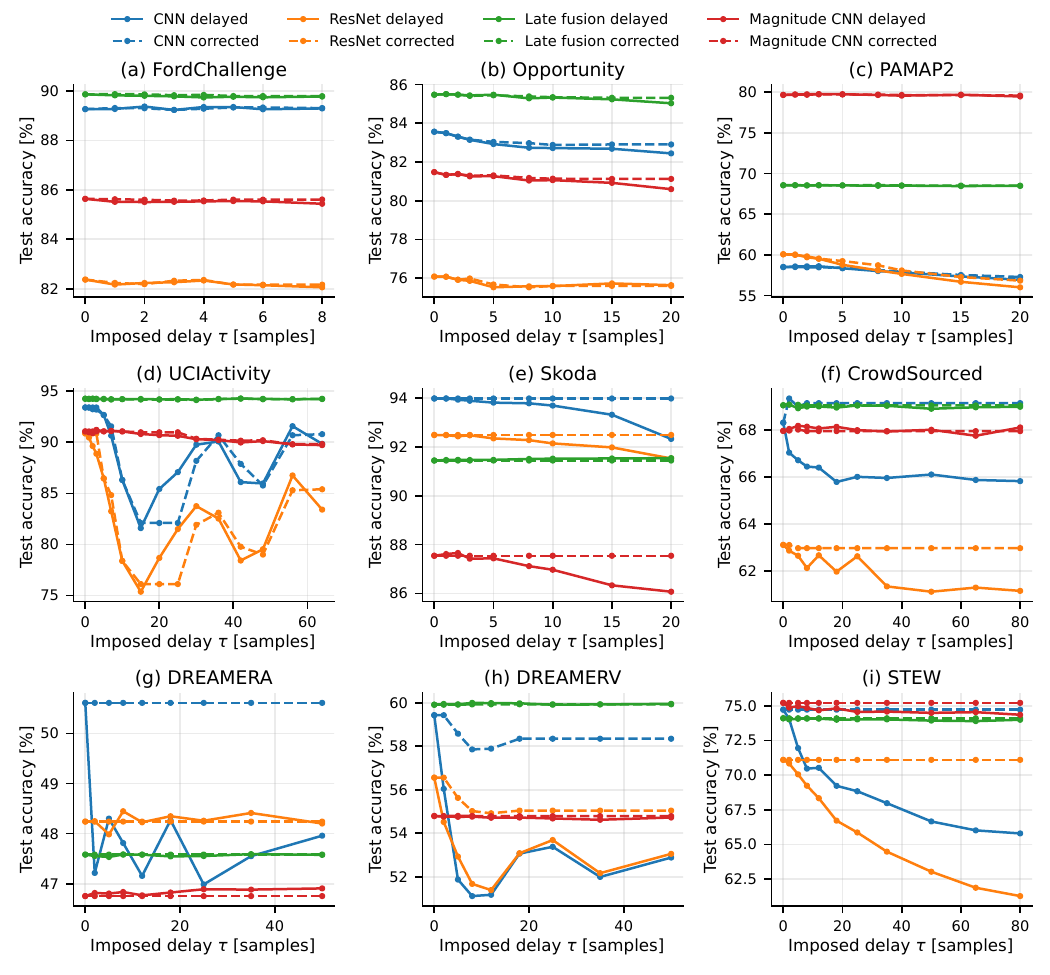}
    \caption{Accuracy of the four classifiers on 9 real-world datasets with imposed lags with and without correction. The solid lines show the average accuracy of the model across the five folds over imposed lags on the test data. The dashed line shows the accuracy after correcting the imposed lag via the alignment loss. For visual clarity, uncertainty bands are not displayed. See appendix for uncertainty bands.}
    \label{fig:realworld_datasets}
\end{figure*}

In this experiment we test how our metric predicts accuracy degradation of classifiers under lag during deployment. To do so, classifiers see only the nominal windows ($\tau=0$) during training. At test time, we delay every sensor group while the complement remains fixed. Note the goal is not to maximize accuracy on any dataset, but to study how misalignment affects performance on different datasets and when correction via our metric is useful. Hence, we are mostly interested in the progression of accuracies as opposed to single values. 

Figure~\ref{fig:realworld_datasets} shows the average accuracy of the four classifiers on 9 real-world datasets across 5 folds with and without correction via our alignment metric. For visual clarity, uncertainty bands are not displayed here, as on some datasets accuracies vary greatly between folds. See appendix for uncertainty bands. We identify three distinct behaviors. On FordChallenge, Opportunity, and PAMAP2, there seems to be only a minor dependence on cross-channel interactions, and hence most classifiers are inherently robust against misalignment. Only CNN and ResNet show some minor dependence on Opportunity and PAMAP2 with larger lags so that our alignment metric can recover some accuracy. UCIActivity is the second group that shows erratic behavior. Here, Late Fusion and Magnitude CNN are stable to lags, whereas CNN and ResNet suffer a sharp performance drop already with smaller lags. Interestingly, these are recovered with larger lags again, but no stable behavior emerges. Consequently, our metric is not able to fully recover performance on this dataset. The last and by far largest group, composed of Skoda, CrowdSourced, DreamerA, DreamerV, and STEW, shows a clear trend towards large performance losses when lags are introduced. For example, on STEW, performance drops from close to 75 \% to below 67 \% for the CNN. On these datasets, our metric is clearly helpful as it recovers the original performance in many cases (e.g., on Skoda, DreamerA, and STEW) or helps to stabilize it (e.g., DreamerV, CrowdSourced).  

As an additional comparison of test-time correction, we adopt SyncNet~\cite{ChungZ16a} which is trained so that embedding distance corresponds to temporal misalignment. We use SyncNet as our state-of-the-art learned synchronizer and leave an exhaustive comparison of signal-level alignment methods for future work. Note, that SyncNet assumes \emph{already aligned} training data and predicts drift from that reference, whereas our metric can prefer a nonzero training alignment as shown in Table~\ref{tab:global-alignment}, so we measure (mean) absolute residual error relative to each method's preferred training-fold baseline. Zero means that the test data were returned to that baseline. This setting favors the learned detector, which receives aligned training pairs and places our metric at a disadvantage. 

\begin{table}[h]
  \centering
  \small
  \caption{Baseline-relative lag correction using the label-free MDL diagnostic and learned SyncNet baseline. Columns report mean absolute residual error normalized by maximum window length without lag. Higher is better.}
  \label{tab:syncnet-comparison}
\begin{tabular}{@{}lrrr@{}}
\toprule
Dataset & Our metric {[}\%{]} & SyncNet {[}\%{]} & Diff {[}pp{]} \\ \midrule
FordChallenge & 93.66 & 94.41 & -0.75 \\
DREAMERV & 98.95 & 100.0 & -1.05 \\
Opportunity & 94.38 & 100.0 & -5.62 \\
UCIActivity & 94.38 & 100.0 & -5.62 \\
PAMAP2 & 89.44 & 100.0 & -10.56 \\ 
\vspace{0.2cm}
CrowdSourced & 55.11 & 78.07 & -22.96 \\
Skoda & 100.0 & 100.0 & 0 \\
\vspace{0.2cm}
STEW & 100.0 & 100.0 & 0 \\
DREAMERA & 100.0 & 99.43 & 0.57 \\
SyncSine-Robust & 68.51 & 65.54 & 2.97\\
SyncSine-Sensitive & 91.47 & 76.05 & 15.42 \\\bottomrule
\end{tabular}
\end{table}

Table \ref{tab:syncnet-comparison} shows the results of the comparison against SyncNet. 
For an easier comparison, we normalize the MAE by the length of each window minus the maximum lag. For example, a value of 95\% means we found the correct lag within 5\% of the window's length. The last column shows the difference in percentage points between MDL and SyncNet, where a negative indicates a better performance by SyncNet and a positive value better performance by MDL.
We see three distinct groups of datasets. In the top group, SyncNet wins over our metric on 6 datasets. The middle group shows a perfect match, where both methods have the same, perfect, performance. In the last group, MDL wins over SyncNet on 3 datasets. We summarize these results as follows: While SyncNet wins on 6 out of 11 datasets, MDL nearly matches SyncNet's performance on four datasets (FordChallenge, DreamerV, Opportunity, UCIAcitivy) and is only substantially outperformed on two datasets (PAMAP2, CrowdSourced). Conversely, on DreamerA and SyncSine-Robust, SyncNet also closely matches MDL's performance and is only outperformed on SyncSine-Sensitive. We conclude that despite the favorable setting for SyncNet in this evaluation, our MDL metric closely follows its performance, showing competitive behavior.

\section{Conclusion}

We introduced a classifier-, label-, and training-free MDL diagnostic for shared temporal structure in multichannel time series. Its central intuition is that one group can be described through a compact representation of the remaining channels when they share aligned temporal structure, and a shift that destroys this structure increases the residual description. The experiments show that the effect of misalignment depends strongly on the task and classifier, and that the lag curve can guide the correction of temporal misalignment. The synthetic examples distinguish alignment-sensitive and robust tasks and recover the expected sine periodicity. On real data, correction often restores or stabilizes performance after an imposed shift, whereas correcting candidate offsets in already processed benchmarks seldom improves accuracy. Unlike a detector tied to a trusted baseline, the metric can also audit that baseline itself. Its nonzero optima on four real datasets warrant further investigation of these datasets. The diagnostic is therefore useful for finding and debugging synchronization issues, while its interpretation remains conditional on the chosen code and representation. Further exploration of different codes and shared representations with an in-depth exploration of deployment-lag correction remains future work.


\bibliography{aaai2027}

\section{APPENDIX}

\section{Class-Conditional Alignment Cost}
\label{sec:conditional-metric}

The label-free metric in the main paper measures shared-representation compressibility over all windows jointly. When labels are available, the same construction can instead be applied within each class. Let $Z_{G,a}^{\tau}(X,Y)$ contain the windows of class $a$ after delaying group $G$ by $\tau$. The class-conditional shared-representation gain is
$$
R_G^{\mathrm{cond}}(\tau)
=
\sum_{a\in\mathcal Y}
\left[
\ell_{\mathrm{ind}}\!\left(Z_{G,a}^{\tau}(X,Y)\right)
-
\ell_{\mathrm{sh}}\!\left(Z_{G,a}^{\tau}(X,Y)\right)
\right].
$$
Thus, every class receives its own anchored PCA representation and reconstruction parameters, and the resulting codelength gains are summed. This can isolate class-specific temporal structure, but requires labels and encodes smaller subsets separately.

We use the same raw-bit definition as in the main paper. For the configured non-negative lag grid
$$
\mathcal T_\Delta\subseteq\{\tau\in\mathbb Z\mid0\leq\tau\leq\Delta\},
$$
including zero, define
\begin{align*}
R_G^{\mathrm{cond},\star}(\Delta)
&=\max_{\tau\in\mathcal T_\Delta}R_G^{\mathrm{cond}}(\tau),\\
Q_G^{\mathrm{cond}}(\tau;\Delta)
&=R_G^{\mathrm{cond},\star}(\Delta)-R_G^{\mathrm{cond}}(\tau).
\end{align*}
The cost $Q_G^{\mathrm{cond}}$ is measured in bits, is non-negative, and is zero at a best observed lag. As for the label-free version,
\begin{align*}
A_G^{\mathrm{cond}}(\Delta)
&=\frac{1}{|\mathcal T_\Delta|-1}
\sum_{\tau\in\mathcal T_\Delta\setminus\{0\}}
Q_G^{\mathrm{cond}}(\tau;\Delta),\\
A^{\mathrm{cond}}(\Delta)
&=\max_{G\in\mathcal G}A_G^{\mathrm{cond}}(\Delta).
\end{align*}
The preferred lag is $\tau_G^\star=\arg\min_{\tau}Q_G^{\mathrm{cond}}(\tau;\Delta)$. As before, raw codelengths should be compared across lags for the same dataset, group, and code. They are not normalized percentages and their magnitude is not directly comparable across datasets. For the code, we use exactly the PCA construction specified in the main paper: rank-one PCA on the unshifted complement, consistently oriented components, non-negative reconstruction slopes, delta-entropy coding for the quantized target and representation, raw entropy coding for residual symbols, and 16 bits per reconstruction parameter. 

\subsection{Code Instantiations}
\label{sec:shared-representation-codes}
The above definition leaves open how the anchored representation $\phi$, the reconstruction family $f_\Theta$, and the component codes are instantiated, and specific choices must be tuned to the specific data and classification task at hand. We state three requirements on the code for a successful adoption. First, the representation must be anchored in $X_{\bar G}$ and must not be estimated from the shifted group $X_G^\tau$. Second, the representation and reconstruction model must be sufficiently simple; otherwise, the code could reconstruct each channel independently and would no longer measure shared temporal structure. Third, the model must be fixed before evaluating lags, so that changes in description length are caused by temporal shifts rather than by changes in the coding scheme.

For the remainder of the paper, we use deliberately simple choices that work well across all of your experiments: In order to compute the bit length of the code, we use a fixed precision. 

For a given float representation, we therefore, for each channel, compute its mean $\mu_c$ and standard deviation $\sigma_c$ over the supplied fold and quantize
$$
\widetilde X_{i,c,t}
=
\operatorname{round}\!\left(
64\,\frac{X_{i,c,t}-\mu_c}{\sigma_c}
\right).
$$
These channel statistics, and the quantization scale is fixed before evaluating any channel group or lag, where the scale of 64 corresponds to a quantization resolution of $\sigma_c/64$. To exploit temporal smoothness, we apply a first-order delta transform to each quantized window. We treat the first value of every window and all subsequent within-window differences as two separate integer-symbol streams. Combining with a plug-in empirical-entropy codelength (c.f. \cite{CoverThomas2006}), the resulting temporal codelength is
$$
 \scalebox{0.9}{$
DL_{\mathrm{diff}}(U)
=
L_{\mathrm{emp}}\left(\{U_{i,c,1}\}_{i,c}\right) +
L_{\mathrm{emp}}\left(\{U_{i,c,t}-U_{i,c,t-1}\}_{i,c,t>1}\right)
$}
$$


For the representation estimator $\phi$, we use rank-one PCA on the unshifted complement. The complement $X_{\bar G}$ is reshaped by stacking samples and time points into a matrix $ X_{\bar G}\in\mathbb R^{N\times C_{\bar G}},$ where $N$ combines the sample and time dimensions. We then compute the first principal component score
$$
S_G=\phi_{\mathrm{PCA}}(X_{\bar G})\in\mathbb R^{N\times 1}.
$$
If the complement contains only one channel, this reduces to using that channel itself as the anchored representation. Otherwise, $S_G$ is the dominant one-dimensional temporal component of the complement. The reconstruction family is linear. For each lag $\tau$, the shifted group is reconstructed as
$$
\hat X_G^\tau
=
S_G W_G^\top+\mathbf 1\mu_G^\top,
$$
where $\Theta_G^\tau=(W_G,\mu_G)$ are fitted for the shifted group at lag $\tau$, subject to $W_G\geq0$, while the offset $\mu_G$ remains unrestricted. The non-negative source weights prevent an anti-phase signal from being treated as aligned merely by negating the shared component. We fix the otherwise arbitrary sign of each PCA component by orienting its largest-magnitude loading positively. This deliberately limited reconstruction model tests whether the shifted group is predictable from a consistently oriented temporal component extracted from the complement.

We encode the anchored representation $S_G$ with the delta entropy code $DL_{diff}$: the first quantized symbol and the within-window temporal differences are encoded using empirical entropy. The target group is encoded directly with the same temporal code. The reconstruction residual $X_G^\tau-\hat X_G^\tau$
is rounded to integer symbols and encoded using empirical entropy without delta encoding. Each reconstruction slope and intercept is assigned 16 bits. The source cost, parameter cost, and residual cost are all included in $\ell_{\mathrm{sh}}$. These choices are held fixed across the complete lag sweep.

\section{Additional Real-World Results}

For reference, we include a more detailed explanation of our experiments here: All results use five folds. The real-world folds are the predefined MONSTER\_TS splits and the synthetic data use five stratified folds. Classifiers are trained only on nominal windows ($\tau=0$). Table \ref{tab:appendix-training} summarizes the training protocol of the four classifiers and our MDL metric. 

\begin{table}
\centering
\scriptsize
\caption{Training and evaluation settings used for every dataset and fold.}
\label{tab:appendix-training}
\begin{tabular}{@{}ll@{}}
\toprule
Setting & Value \\
\midrule
Random seed & $42+{}$fold index \\
Validation split & stratified 15\% of training fold \\
Input normalization & per-channel training mean/std. \\
Training / evaluation batch & 256 / 2048 \\
Maximum epochs / early stopping & 100 / patience 10 \\
Classifier optimizer & Adam, $10^{-4}$ \\
Classifier LR schedule & factor $0.5$, patience 3 \\
SyncNet optimizer & AdamW, $10^{-4}$, wd. $5\cdot10^{-4}$ \\
SyncNet contrastive margin & 1 \\
Arithmetic precision & 32 bit \\
Metric quantization & $64$ levels per training std. \\
PCA rank / parameter cost & 1 / 16 bits per parameter \\
\bottomrule
\end{tabular}
\end{table}

At evaluation time, each sensor group is delayed in turn while the complement remains fixed, and windows are cropped to their common valid region rather than padded. For correction and the SyncNet comparison, we use the group with the largest worst-lag accuracy loss averaged over the four classifiers and five folds. The unconditional MDL correction is computed directly on the unlabeled test windows. For an imposed lag $t$, it searches feasible corrections that move the residual lag toward zero and selects the residual lag with minimum raw alignment cost. Table~\ref{tab:appendix-lags} shows the evaluated lags with the selected worst-case group. For the synthetic datasets we evaluate lags from 0 to 1500 in increments of twenty. 

\begin{table}
\centering
\scriptsize
\caption{Configured non-negative lag grids and the sensor group used for correction and the SyncNet comparison. The notation $a{:}s{:}b$ denotes values from $a$ to $b$ in steps of $s$.}
\label{tab:appendix-lags}
\begin{tabular}{@{}p{0.25\columnwidth}p{0.48\columnwidth}p{0.17\columnwidth}@{}}
\toprule
Dataset & Evaluated lags $\mathcal T_\Delta$ [samples] & Selected group \\
\midrule
FordChallenge & 0, 1, 2, 3, 4, 5, 6, 8 & vehicle \\
Opportunity & 0, 1, 2, 3, 5, 8, 10, 15, 20 & l shoe \\
PAMAP2 & 0, 1, 2, 3, 5, 8, 10, 15, 20 & ankle \\
UCIActivity & 0, 1, 2, 3, 5, 7, 10, 15, 20, 25, 30, 36, 42, 48, 56, 64 & gyro \\
Skoda & 0, 1, 2, 3, 5, 8, 10, 15, 20 & s06 \\
CrowdSourced & 0, 2, 5, 8, 12, 18, 25, 35, 50, 65, 80 & F7 \\
DREAMERA / DREAMERV & 0, 2, 5, 8, 12, 18, 25, 35, 50 & ch 12 / ch 13 \\
STEW & 0, 2, 5, 8, 12, 18, 25, 35, 50, 65, 80 & ch 12 \\
SyncSine-Sensitive / Robust & $0{:}20{:}1500$ & ch1 / ch0 \\
\bottomrule
\end{tabular}
\end{table}

\begin{figure*}[!p]
\centering
\includegraphics[width=0.99\textwidth]{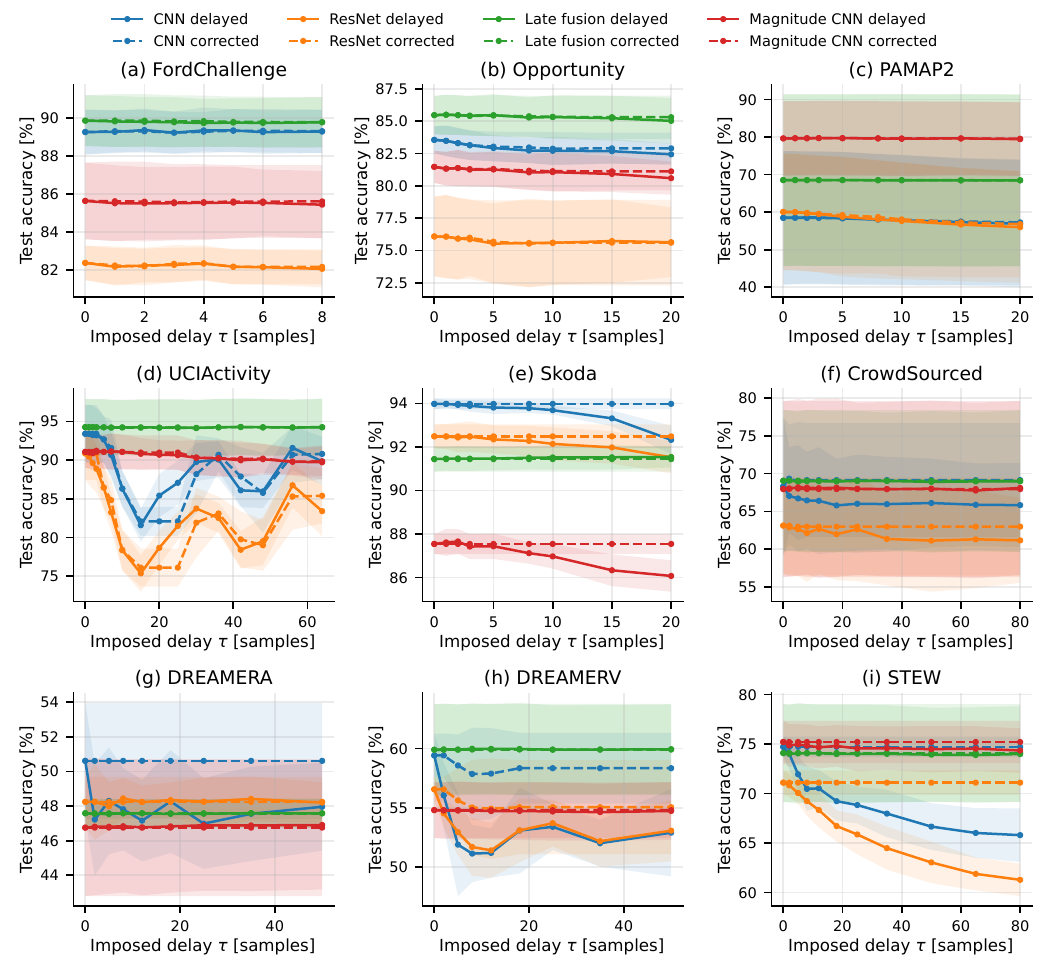}
\caption{Test accuracy under imposed sensor delays before and after label-free MDL correction. Lines show fold means and bands show one standard deviation over five folds. Solid lines are evaluated at the imposed lag. Dashed lines use the residual lag selected by the MDL alignment cost.}
\label{fig:realworld-datasets-uncertainty}
\end{figure*}

Figure~\ref{fig:realworld-datasets-uncertainty} supplements the correction figure in the main paper with one-standard-deviation bands over folds. The wide bands on several datasets reflect differences between folds rather than uncertainty about individual predictions; the mean curves are identical to those shown in the main paper.

\begin{figure*}[!p]
\centering
\includegraphics[width=0.99\textwidth]{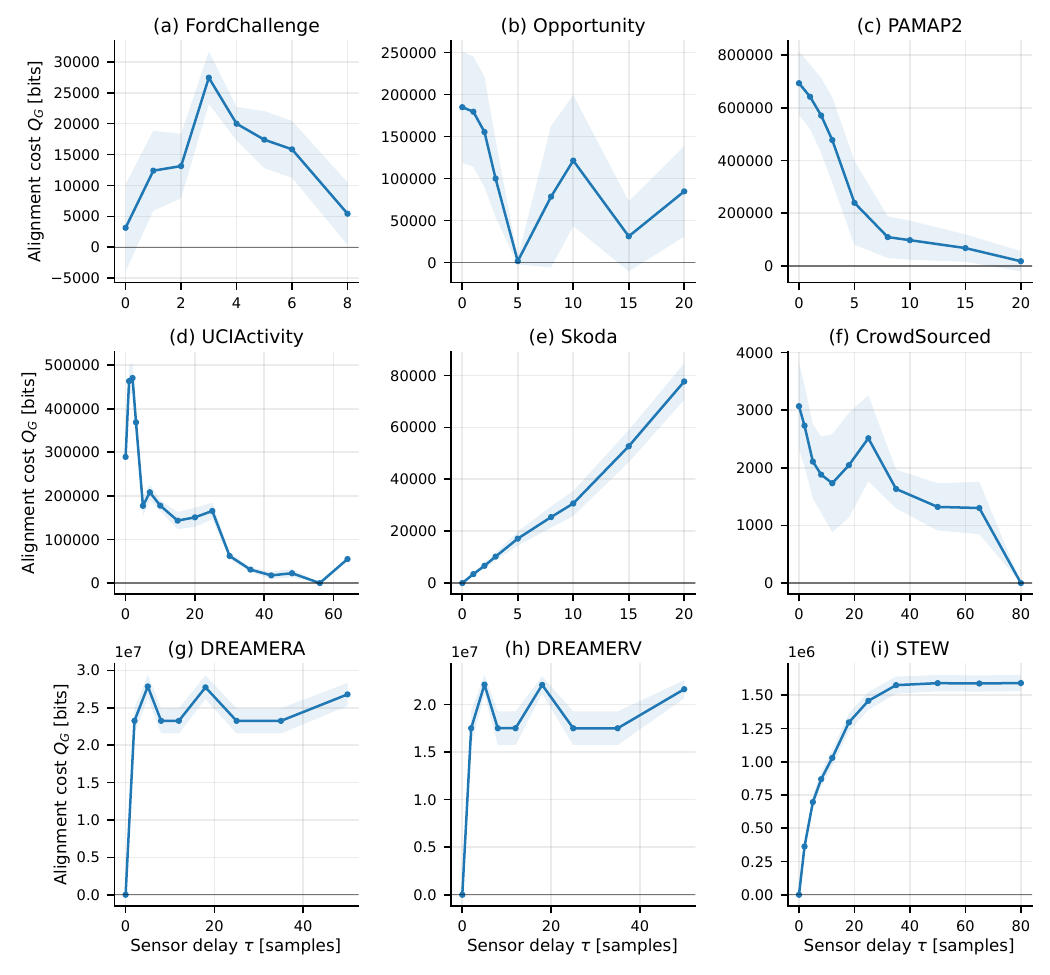}
\caption{Label-free alignment cost over all evaluated real-world datasets. Lines show fold means and bands show one standard deviation. Each panel uses the classifier-selected group from Table~\ref{tab:appendix-lags}. Raw bit scales are dataset-specific.}
\label{fig:realworld-metric-all}
\end{figure*}

Figure~\ref{fig:realworld-metric-all} shows the complete real-world alignment-cost grid. FordChallenge, Opportunity, PAMAP2, and UCIActivity attain a lower cost at a non-zero lag for the selected group, whereas Skoda, CrowdSourced, DREAMERA, DREAMERV, and STEW are minimized at or close to zero in the whole-dataset audit. The fold bands additionally show where the estimated curve is stable across data splits.

\section{Complete SyncNet Comparison}

Tables~\ref{tab:appendix-syncnet-error} and~\ref{tab:appendix-syncnet-accuracy} extend the lag-error comparison in the main paper with the resulting classifier accuracies. Both methods are evaluated on the same group, folds, and common lag grid. Following the main-paper protocol, only imposed lags for which both methods' preferred training baselines are reachable are included. Lag error is measured relative to each method's own preferred training baseline. Accuracy is obtained from the saved lag curves of the four classifiers: for every selected residual lag, we evaluate the corresponding already-computed classifier accuracy, then average over classifiers, folds, and included imposed lags. Thus, the tables add no model evaluations.

SyncNet more accurately returns most real-world datasets to its learned baseline, which is reflected by the larger recovery on PAMAP2 and UCIActivity. MDL is competitive on the zero-baseline datasets and substantially reduces the synthetic sensitive task's accuracy loss despite its periodic ambiguity. CrowdSourced and the robust synthetic task have boundary or weakly identified MDL optima, so their lag errors should not be interpreted as localized estimates. Most importantly, the comparison concerns deployment drift relative to a learned baseline; only the MDL criterion can also question that baseline without training.

\begin{table}
\centering
\scriptsize
\caption{Baseline-relative lag errors in samples (lower is better). Cases counts the jointly reachable fold--lag pairs.}
\label{tab:appendix-syncnet-error}
\begin{tabular}{@{}lrrr@{}}
\toprule
Dataset & MDL & SyncNet & Cases \\
\midrule
FordChallenge & 2.03 & 1.79 & 33 \\
Opportunity & 4.50 & 0.00 & 22 \\
PAMAP2 & 8.44 & 0.00 & 9 \\
UCIActivity & 3.60 & 0.00 & 10 \\
Skoda & 0.00 & 0.00 & 45 \\
CrowdSourced & 79.00 & 38.60 & 5 \\
DREAMERA & 0.00 & 1.18 & 45 \\
DREAMERV & 2.16 & 0.00 & 45 \\
STEW & 0.00 & 0.00 & 55 \\
\midrule
SyncSine-Sensitive & 128.00 & 359.18 & 220 \\
SyncSine-Robust & 472.29 & 516.86 & 70 \\
\bottomrule
\end{tabular}
\end{table}

\begin{table}
\centering
\scriptsize
\caption{Classifier accuracy after lag correction [\%]. Values average CNN, ResNet, late fusion, and Magnitude CNN over five folds and the jointly reachable imposed lags.}
\label{tab:appendix-syncnet-accuracy}
\begin{tabular}{@{}lrrrr@{}}
\toprule
Dataset & Nominal & Lagged & MDL & SyncNet \\
\midrule
FordChallenge & 86.5 & 86.5 & 86.5 & 86.5 \\
Opportunity & 81.6 & 81.1 & 81.2 & 81.6 \\
PAMAP2 & 74.0 & 72.6 & 72.8 & 74.0 \\
UCIActivity & 92.4 & 89.9 & 89.9 & 92.4 \\
Skoda & 91.4 & 91.1 & 91.4 & 91.4 \\
CrowdSourced & 67.1 & 66.0 & 67.3 & 66.3 \\
DREAMERA & 48.3 & 47.7 & 48.3 & 48.3 \\
DREAMERV & 57.7 & 55.3 & 57.2 & 57.7 \\
STEW & 73.8 & 71.2 & 73.8 & 73.8 \\
\midrule
SyncSine-Sensitive & 91.5 & 78.2 & 90.6 & 91.2 \\
SyncSine-Robust & 100.0 & 100.0 & 100.0 & 100.0 \\
\bottomrule
\end{tabular}
\end{table}

\section{Class-Conditional versus Label-Free Metric}

Tables~\ref{tab:conditional-lags} and~\ref{tab:conditional-costs} compare both variants on the same classifier-selected group and lag grid. The reported lag is the median fold optimum with its fold range, and $A_G$ is the mean raw alignment cost over non-zero lags, averaged over folds. The bit values are descriptive within each row and variant; class-conditional coding changes the symbol populations and therefore need not preserve the unconditional scale.

\begin{table}
\centering
\scriptsize
\caption{Preferred lags for the label-free (U) and class-conditional (C) metrics, reported as median [minimum--maximum] over five folds.}
\label{tab:conditional-lags}
\begin{tabular}{@{}llrr@{}}
\toprule
Dataset & Group & U lag & C lag \\
\midrule
FordChallenge & vehicle & 0 [0--8] & 6 [0--8] \\
Opportunity & l shoe & 5 [5--15] & 15 [0--15] \\
PAMAP2 & ankle & 20 [8--20] & 3 [0--5] \\
UCIActivity & gyro & 56 [56--56] & 64 [64--64] \\
Skoda & s06 & 0 [0--0] & 0 [0--0] \\
CrowdSourced & F7 & 80 [80--80] & 65 [65--80] \\
DREAMERA & ch 12 & 0 [0--0] & 0 [0--0] \\
DREAMERV & ch 13 & 0 [0--0] & 0 [0--0] \\
STEW & ch 12 & 0 [0--0] & 0 [0--0] \\
\midrule
SyncSine-Sensitive & ch1 & 640 [640--640] & 0 [0--640] \\
SyncSine-Robust & ch0 & 1300 [760--1480] & 1480 [760--1500] \\
\bottomrule
\end{tabular}
\end{table}

\begin{table}
\centering
\scriptsize
\caption{Mean raw alignment cost $A_G$ [kbit] for the label-free (U) and class-conditional (C) metrics.}
\label{tab:conditional-costs}
\begin{tabular}{@{}lrr@{}}
\toprule
Dataset & U $A_G$ & C $A_G$ \\
\midrule
FordChallenge & 15.9 & 17.2 \\
Opportunity & 94.1 & 122.2 \\
PAMAP2 & 278.1 & 51.7 \\
UCIActivity & 167.5 & 166.1 \\
Skoda & 28.0 & 82.8 \\
CrowdSourced & 1.7 & 25.2 \\
DREAMERA & 24834.9 & 24824.0 \\
DREAMERV & 19162.3 & 18970.9 \\
STEW & 1206.3 & 1230.8 \\
\midrule
SyncSine-Sensitive & 36.2 & 72.5 \\
SyncSine-Robust & 0.5 & 0.7 \\
\bottomrule
\end{tabular}
\end{table}

The variants agree on the clear zero-lag optima of Skoda, DREAMERA, DREAMERV, and STEW and identify similar broad non-zero structure for UCIActivity. On the periodic sensitive task, zero and approximately one period are equivalent alignments, so the different discrete optimum does not constitute a contradiction. Neither variant localizes the robust synthetic task: both costs are tiny and the fold optima span most of the grid. On the remaining real-world datasets, conditioning changes the preferred lag without producing a consistently more stable estimate. Since it offers no systematic empirical advantage and requires labels, we use the unconditional metric throughout the main paper.

\end{document}